\documentclass[conference]{IEEEtran}
\usepackage{amsmath,amssymb}
\usepackage{psfrag}
\usepackage{epsfig}
\usepackage{cite}
\usepackage{graphics}
\usepackage{color}
\usepackage{subfigure}
\usepackage{multirow}
\usepackage{threeparttable}%%%LBL
\usepackage{booktabs}%%%LBL
\usepackage{mathrsfs}
\usepackage{amsfonts}
\usepackage{psfrag}
\usepackage{algorithmic}
\usepackage{algorithm}
\usepackage{cases}

\usepackage{multirow}
\usepackage[left=0.71in,top=0.94in,right=0.71in,bottom=1.18in]{geometry}
\begin{document}
\title{ PSA: A Novel Optimization Algorithm Based on Survival Rules of Porcellio  Scaber}
\author{\IEEEauthorblockN{Yinyan Zhang}
\IEEEauthorblockA{College of Cyber Security\\Jinan University\\
Guangzhou, China\\
Email: yyzhang@jnu.edu.cn} \and \IEEEauthorblockN{Pei Zhang}
\IEEEauthorblockA{College of Cyber Security\\Jinan University\\
Guangzhou, China\\
Email: 1207382268@qq.com} \and \IEEEauthorblockN{Shuai Li}
\IEEEauthorblockA{College of Engineering\\Swansea University\\
Swansea SA1 7EN, U.K.\\
Email: shuaili@ieee.org} }

%\author{Yinyan Zhang, Pei Zhang, Shuai Li
%\thanks{ This work is supported by the Fundamental Research
%Funds for the Central Universities under Grant 21620346 and the
%National Natural Science Foundation of China under Grant 61932011.}
%\thanks{Y. Zhang and P. Zhang are with the College of Cyber Security, Jinan University,
%Guangzhou 510632, China (e-mails: yyzhang@jnu.edu.cn (Y. Zhang);
%1207382268@qq.com (P. Zhang)).}
%\thanks{S. Li is with the College of Engineering, Swansea University,
%Swansea SA1 7EN, U.K. (e-mail: shuaili@ieee.org).
%}% <-this % stops a space
%}

\maketitle
\begin{abstract}
Bio-inspired algorithms such as neural network algorithms and
genetic algorithms have received a significant amount of attention
in both academic and engineering societies. In this paper, based on
the observation of two major survival rules of a species of
woodlice, i.e., porcellio scaber, we present an algorithm called the
porcellio scaber algorithm (PSA) for solving general unconstrained
optimization problems, including differentiable and non-differential
ones as well as the case with local optima. Numerical results based
on benchmark problems are presented to validate the efficacy of PSA.
\end{abstract}

% key words

\IEEEpeerreviewmaketitle

\section{Introduction}

 In recent decades, bio-inspired algorithms, such as neural networks,
genetic algorithms \cite{ai}, are widely investigated and applied in
the fields of pattern recognition \cite{nc}, robotics \cite{long},
control \cite{chen}, task scheduling \cite{ad}, and optimization
\cite{ssg}. Compared with traditional methods, bio-inspired
algorithms are more efficient in handling complicated optimization
problems, such as the NP-hard traveling salesman problem.  Kennedy
and Eberhart \cite{1} proposed particle swarm optimization, which is
one of the well known bio-inspired algorithms. It is based on the
swarm behavior of animals, such as fish and bird schooling. Inspired
by how ants find a shortest path, Dorigo and Gambardella \cite{2}
proposed an ant colony system algorithm for solving the traveling
salesman problem. Based on some idealized behaviors of the flashing
characteristics of fireflies, Yang {\it et al.} \cite{3} proposed a
firefly algorithm for solving non-convex economic dispatch problems
with valve loading effect. Inspired by the competitive and
self-interested behaviors of political party individuals, Huan {\it
et al.} \cite{aa1} proposed a so-called ideology algorithm for
solving unconstrained optimization problems. The performance of
these algorithms indicates that bio-inspired algorithms can be
efficient and promising.

In this paper, based on the observation of two survival rules of a
species of woodlice, i.e., porcellio scaber (PS) \cite{5,sur,6}, we
present a novel algorithm called the porcellio scaber algorithm
(PSA) for solving general unconstrained optimization problems
including differentiable and non-differentiable ones as well as the
case with local optima. The performance of the proposed algorithm is
evaluated via benchmark problems.

\section{Porcellio  scaber-inspired algorithm design}\label{sec.2}

\begin{figure}[t]\centering
\includegraphics[scale=0.1]{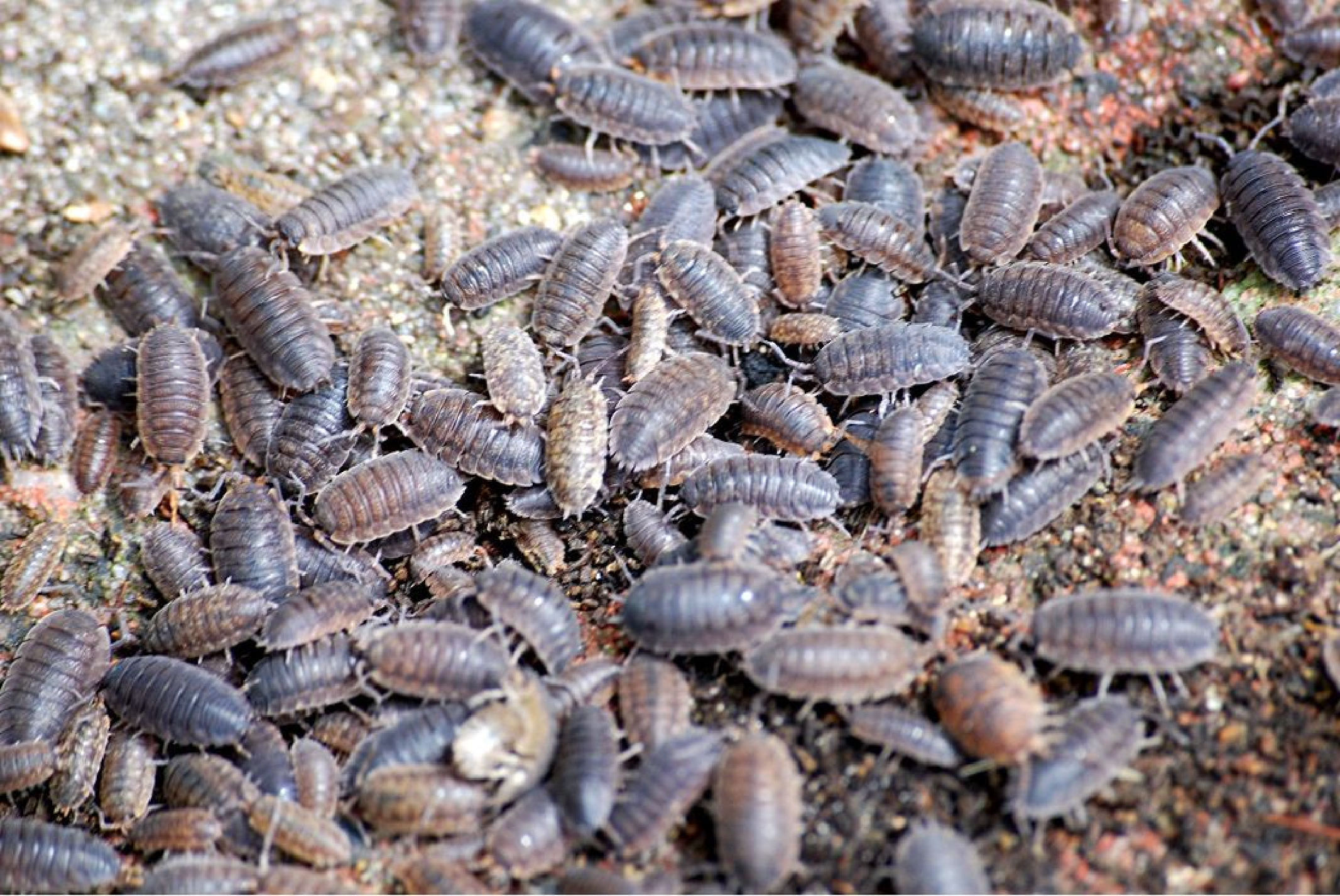}
\caption{Porcellio scaber by Peter R\"uhr, used
 under the Creative Commons Attribution 3.0 Unported license
\cite{4}.}\label{fig.1}
\end{figure}

Porcellio scaber, as shown in Fig. \ref{fig.1} is a species of
woodlice. They prefer to live in moist, dark, and cool places and
are known to live in groups \cite{5}. Due to the outstanding
behavioral advantage of porcellio scaber, some of its varieties even
can survive in extremely harsh environments, e.g., deserts around
North Africa and the Middle East, where the searching of moist,
dark, and cool places is usually difficult \cite{sur}. To avoid
desiccation and a rapid drop in temperature, both of which could
cause significant mortality, they shelter. In addition to their
eyes, porcellio scaber have many sensory receptors on their body,
making it possible for them to detect chemical, mechanical, and
hygrometric conditions about the environment \cite{6}. When the
environment conditions are not favourable, porcellio scaber keep
moving until morality or reach places with good conditions. Besides,
for worse environment conditions, they move more quickly. On the
other hand,
 when the environment condition is optimal, i.e., sufficiently good,
they stay there.

 For the sake of illustration, let the position of the
mass center of each porcellio scaber at the $k$th ($k=0,1,2,\cdots$)
time instant be denoted by a vector $\mathbf{x}^k_i$ and let the
environmental condition at a position $\mathbf{x}$ by described by a
fitness function $f(\mathbf{x})$, where a minimum value of
$f(\mathbf{x})$ corresponds to the optimal environmental condition
for porcellio scaber.
 We consider the case that
there is a group of totally $N$ porcellio scaber, i.e.,
$i\in\{1,2,\cdots,N\}$.

\begin{figure}[t]\centering
\subfigure[]{\includegraphics[scale=0.25]{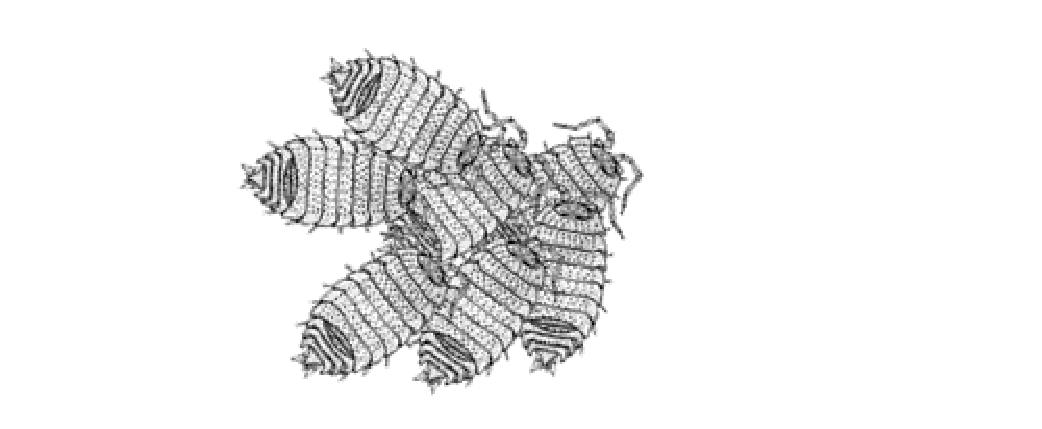}}
\subfigure[]{\includegraphics[scale=0.25]{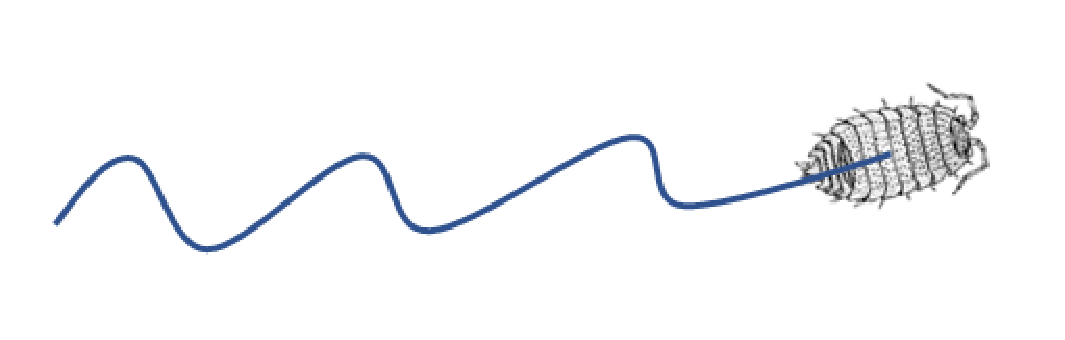}} \caption{Two
behaviors of porcellio scaber. (a) Group behavior: aggregation. (b)
Individual behavior: propensity to explore novel environments.
}\label{fig.2}
\end{figure}

\begin{algorithm}[t]
\caption{PSA}
\begin{algorithmic}
\STATE {\it Objective function} $f(\mathbf{x})$,
~$\mathbf{x}=[x_1,x_2,\cdots,x_d]^\text{T}$ \STATE {\it Generate
initial position of porcellio scaber}
$\mathbf{x}^0_i~(i=1,2,\cdots,N)$ \STATE {\it Environment condition}
$E_{\mathbf{x}}$ {\it at position} $\mathbf{x}$ {\it is determined
by} $f(\mathbf{x})$ \STATE {\it Set weighted parameter} $\lambda$
{\it for decision based on aggregation and the propensity to explore
novel environments} \WHILE{$k<MaxStep$} \STATE {\it Get the position
with the best environment condition, i.e.,}
$\mathbf{x}_*=\text{arg}\min_{\mathbf{x}^k_j}\{f(\mathbf{x}^k_j)\}$
{\it at the current time among the group of porcellio scaber} \STATE
{\it Randomly chose a direction
$\tau=[\tau_1,\tau_2,\cdots,\tau_d]^\text{T}$ to detect} \STATE {\it
Detect the best environment condition} $\min\{E_{\mathbf{x}}\}$ {\it
and worst environment condition} $\max\{E_{\mathbf{x}}\}$ {\it at
position} $\mathbf{x}^{k}_i+\tau$ {\it for} $i=1:N$ {\it all} $N$
{\it porcellio scaber} \FOR{$i=1:N$~{\it all $N$ porcellio scaber}}
\STATE {\it Determine the difference with respect to the position to
aggregate i.e.,}
$\mathbf{x}^k_i-\text{arg}\min_{\mathbf{x}^k_j}\{f(\mathbf{x}^k_j)\})$
 \STATE {\it Determine where to explore, i.e.}, $p\tau$
 \STATE {\it Move to a new position according to the weighted decision
 between aggregation and propensity to explore novel
environments} \ENDFOR \ENDWHILE \STATE{\it Output $\mathbf{x}_*$ and
the corresponding function value $f(\mathbf{x}_*)$} \STATE {\it
Visualization}
\end{algorithmic}
\end{algorithm}

Many studies reveal that porcellio scaber have two behaviors, which
are viewed as their survival rules. The two behaviors are depicted
in Fig. \ref{fig.2}. One is called aggregation and the other is
called the propensity to explore novel environments \cite{5,6,7}.
Note that they aggregate at the places with good environment
conditions \cite{7}. To model the aggregation behavior, we can
describe the movement of porcellio scaber as follows:
\begin{equation}\label{eq.1}
\mathbf{x}^{k+1}_i=\mathbf{x}^{k}_i-(\mathbf{x}^k_i-\text{arg}\min_{\mathbf{x}^k_j}\{f(\mathbf{x}^k_j)\})
\end{equation}
which can be rewritten as
$$\mathbf{x}^{k+1}_i=\text{arg}\min_{\mathbf{x}^k_j}\{f(\mathbf{x}^k_j)\}.$$
Evidently, according to (\ref{eq.1}), each porcellio scaber will
finally stay at the same position  with the best environmental
condition among all their initial positions $\mathbf{x}^{0}_i$. With
only the aggregation behavior, porcellio scaber cannot live in case
that they are initially placed at bad environment. Studies show
that, each porcellio scaber has the propensity to explore novel
environments \cite{6}. Thus, the actual movement of each porcellio
scaber is a weighted result of aggregation and the propensity to
explore novel environments. By considering this behavior, we modify
the model (\ref{eq.1}) and present the following PSA update rule to
solve the problem :
\begin{equation}\label{eq.2}
{\mathbf{x}^{k+1}_i=\mathbf{x}^{k}_i-(1-\lambda)(\mathbf{x}^k_i-\text{arg}\min_{\mathbf{x}^k_j}\{f(\mathbf{x}^k_j)\})-\lambda
p\tau}
\end{equation}
where $p$ is a function to map the fitness of a porcellio scaber to
an action strength. A simple choice is $p=f(\mathbf{x}^k_i+\tau)$.
As verified in our extensive experiments, the following normalized
fitness can achieve better performance:
\begin{equation*}
p=\frac{f(\mathbf{x}^k_i+\tau)-\min\{f(\mathbf{x}^k_i+\tau)\}}{\max\{f(\mathbf{x}^k_i+\tau)\}-\min\{f(\mathbf{x}^k_i+\tau)\}}
\end{equation*}
In the model (\ref{eq.2}), the term $\lambda\in(0,1)$ accounts for
the weight between aggregation and the propensity to explore novel
environments. The value of $\lambda$ can be different for each
different porcellio scaber. The term $p\tau$ corresponds to the
propensity to explore novel environments, where $\tau$ is a random
vector that has the same dimension with that of $\mathbf{x}^k_i$.
Specifically, $p\tau$ means that each porcellio scaber randomly
chooses a direction with respect to their positions of mass centers
to detect the environment condition of surroundings. In addition,
the $p$ term means movement speed requirement which indicates that
the exploration for novel environments is also a result of group
negotiation.

The optimization problem considered in this paper is depicted as
follows:
\begin{equation}\label{pro}
\min_{\mathbf{x}} f(\mathbf{x})
\end{equation}
where function $f$ is lower bounded but can be non-differentiable or
non-convex and can have multiple local minima.
 By the
above observations about the behaviors of porcellio scaber, the PSA
for solving optimization problem (\ref{pro}) is designed and
depicted in Algorithm 1. The main idea is to view the function to be
minimized as an evaluation of the environmental condition from the
perspective of porcellio scaber, and view the decision variable as
the position vector of a porcellio scaber. As the movement result of
porcellio scaber is to stay at a place with the best environment
condition, the algorithm design based on the observation of the
movement behaviors of porcellio scaber is expected to be successful.

\begin{figure}[t]\centering
\includegraphics[scale=0.55]{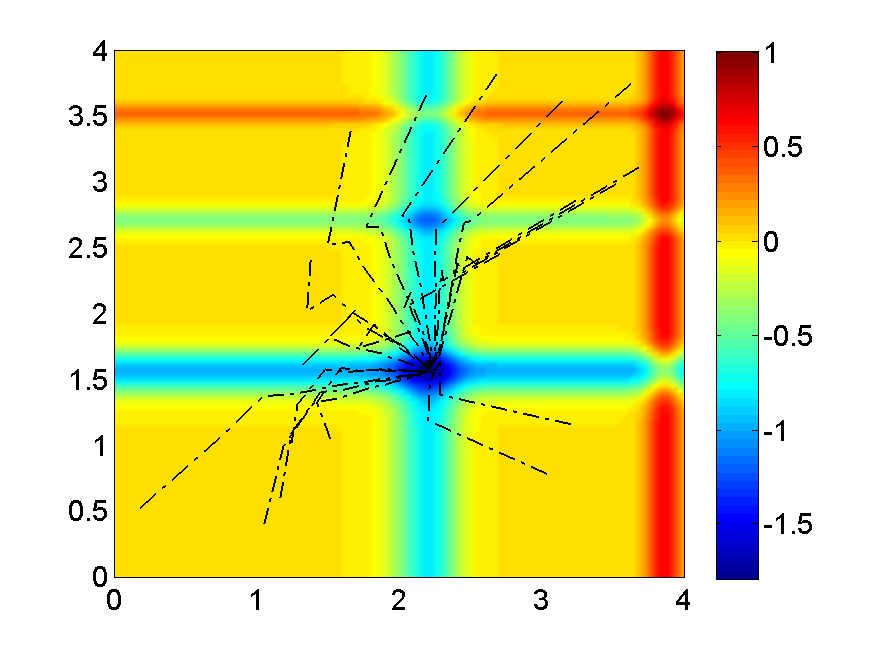}
\caption{Simulation result of the proposed PSA for finding the
minimizer of the Michalewicz function  (\ref{fun.1}) by using 20
porcellio scaber starting from randomly generated initial positions
and by taking 40 steps of movement.}\label{fig.3}
\end{figure}

About the problem considered in this paper, we have the following
remark.

{\it Remark 1:} The unconstrained optimization problem considered in
this paper is general as it can be non-differentiable and
non-convex. It should be noted that the differentiability of the
objective function and the convexity of the problem are often
assumed to be satisfied for existing optimization methods, such as
\cite{Boyd04,Neculai}.

\section{Benchmark validation}\label{sec.3} To test the efficacy of the proposed
PSA, some benchmark problems are used.

\subsection{Example 1}
We consider the Michalewicz function \cite{mic}:
\begin{equation}\label{fun.1}
f(\mathbf{x})=-\sum^d_{i=1}\sin(x_i)[\sin(\frac{ix^2_i}{\pi})]^{2m}
\end{equation}
where $m=10$ and $d=1,2,\cdots$. When $d=2$, the global minimum of
the Michalewicz function is $f_*\approx-1.801$ which occurs at
$\mathbf{x}_*\approx(2.2032,1.5705)$. Under the setting that each
element of $\tau$ is a zero-mean random number with the standard
deviation being 0.001, with $\lambda=0.8$, 20 porcellio scaber, and
40 steps, i.e., $N=20$ and $MaxStep=40$, the best numerical
experiment result is visualized in Fig. \ref{fig.3}. As seen from
this figure, although there are some local minimums, all the
porcellio scaber aggregate at about the optimal one. Numerically,
the solution obtained in the numerical experiment is $\mathbf{x}_*=[
2.202847772916551,1.570778088262819]^\text{T}$ with the
corresponding minimum of the function being
$f(\mathbf{x}_*)=-1.801303342428961\approx-1.801$. The result
validates the efficacy of the proposed method. We have tested the
algorithm for many times, and the successful rate in finding the
global optimizer is relatively high. In addition, when it fails to
find the optimal one, the result is near optimal, i.e., it finds a
local minimum.

\begin{figure}[t]\centering
\includegraphics[scale=0.55]{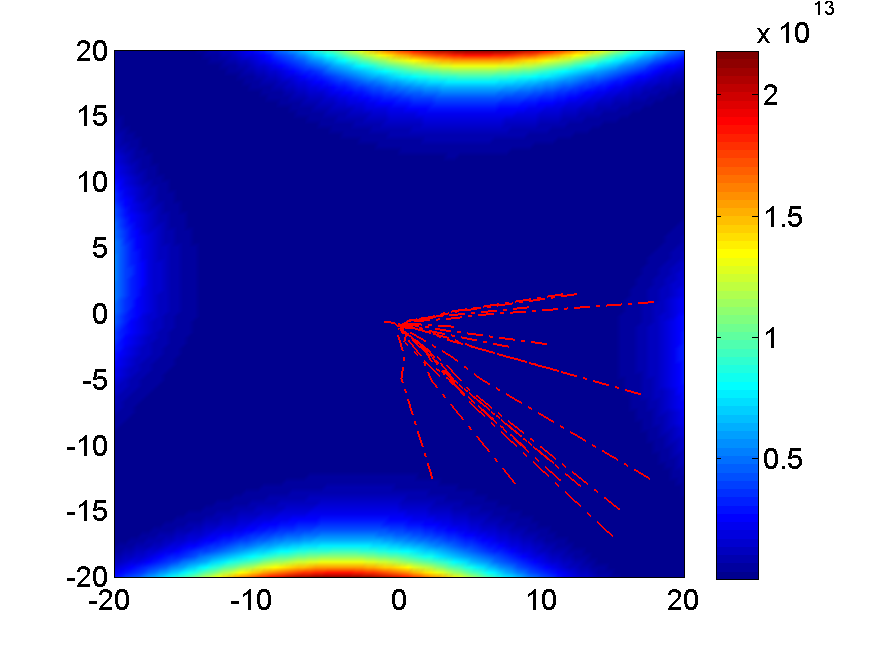}
\caption{Visualization of the numerical result of the proposed PSA
for solving the Goldstein and Price problem (\ref{fun.2}) by using
20 porcellio scaber starting from randomly generated initial
positions and by taking 40 steps of movement.}\label{fig.4}
\end{figure}

\subsection{Example 2}

We also consider the Goldstein and Price problem \cite{gold}:
\begin{equation}\label{fun.2}
\begin{aligned}
\min_{\mathbf{x}}
f(\mathbf{x})=[1+(x_1+x_2+1)^2(19-14x_1+3x^2_1
\\-14x_2+6x_1x_2+3x^2_2)][30+(2x_1-3x_2)^2
\\(18-32x_1+12x^2_1+48x_2-36x_1x_2+27x^2_2)]
\end{aligned}
\end{equation}
The function has four local minima and the global minimizer is
$\mathbf{x}_*=[0,-1]^\text{T}$ with the minimum being $f_*=3$ in the
region established by the constraints $x_1\geq-2$ and $x_2\leq2$. By
using 20 porcellio scaber and taking 40 steps of movement, with
$\lambda=0.6$ and the other parameters set the same as those in the
previous experiment, the best numerical experiment result of the
proposed PSA for solving the Goldstein and Price problem problem
(\ref{fun.2}) is visualized in Fig. \ref{fig.4}. Note that, the
initial positions of the 20 porcellio scaber are randomly set and
are guaranteed to satisfy the inequality constraints $x_1\geq-2$ and
$x_2\leq2$ so as to find the optimal solution in the region. The
solution obtained in the experiment by the algorithm is
$\mathbf{x}_*=[-0.000033275995519,-1.000060284512512]^\text{T}$ with
the corresponding function value being
$f(\mathbf{x}_*)=3.000001415798920\approx3$. The result further
validates the efficacy of the proposed PSA.

\begin{figure}[t]\centering
\includegraphics[scale=0.58]{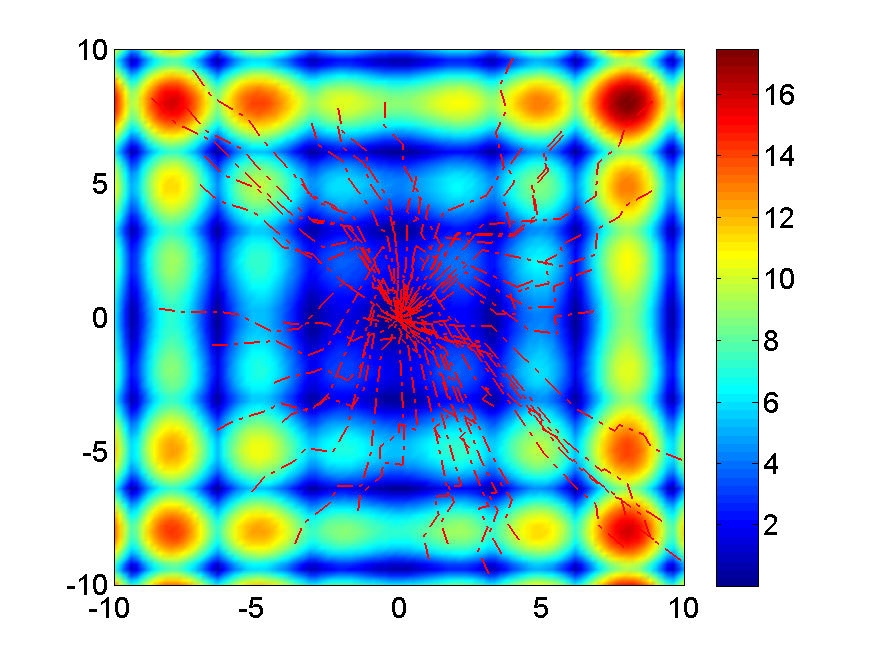}
\caption{Visualization of the numerical result of the proposed PSA
for solving the 2-D Alpine-1-function problem (\ref{fun.3}) by using
50 porcellio scaber starting from randomly generated initial
positions and by taking 100 steps of movement.}\label{fig.5}
\end{figure}

\subsection{Example 3}

We further consider a nondifferentiable optimization problem, i.e.,
the so-called 2-D Alpine-1-function problem \cite{ben}:
\begin{equation}\label{fun.3}
\begin{aligned}
\min_{\mathbf{x}}
f(\mathbf{x})&=\sum^{D=2}_{i=1}|x_i\sin(x_i)+0.1x_i|.
\end{aligned}
\end{equation}
The function has many local minimizers and only one global minimizer
$\mathbf{x}_*=[0,0]^\text{T}$ with the the minimum being $f_*=0$ in
the region established by the constraints $-10\leq x_i\leq10$ with
$i=1,2$. By using 50 porcellio scaber and taking 100 steps of
movement, with $\lambda=0.9$ and the other parameters set the same
as those in the previous experiment, the best numerical experiment
result of the proposed PSA for solving the 2-D Alpine-1-function
problem (\ref{fun.3}) is visualized in Fig. \ref{fig.5}. Note that,
the initial positions of the 50 porcellio scaber are randomly set
and are guaranteed to satisfy the inequality constraints $-10\leq
x_i \leq10$ so as to find the optimal solution in the region. The
solution obtained in the experiment by the algorithm is
$\mathbf{x}_*=[-0.0000066443218359,-0.000001838026950]^\text{T}\approx[0,0]^\text{T}$
with the corresponding function value being
$f(\mathbf{x}_*)=8.478271919968875\times10^{-6}\approx0$. The result
further validates the efficacy of the proposed PSA for the case with
a non-differentiable objective function.

\section{Conclusions}\label{sec.4}

In this paper, a novel bio-inspired algorithm called the PS
algorithm has been  proposed for solving optimization problems. Both
the biology background and implementation of the proposed algorithm
have been illustrated. Numerical experiment results on benchmark
problems have verified the efficacy of the proposed algorithm. The
extension of the algorithm to constrained optimization problems is
illustrated in \cite{psafs}.

% optional entry into table of contents (if used)
\section*{Acknowledgment} This work is
supported by the Fundamental Research Funds for the Central
Universities under Grant 21620346



\end{document}